\documentclass[a4paper]{article}

\usepackage{INTERSPEECH2018}
\usepackage{url}
\usepackage{multirow}
\usepackage{cite}
\usepackage{subfig}
\usepackage{microtype}

\title{ASR-free CNN-DTW keyword spotting using \\ multilingual bottleneck features for almost zero-resource languages} 
\name{Raghav Menon$^1$, Herman Kamper$^1$, Emre Yilmaz$^{2,3}$, John Quinn$^{4,5}$, Thomas Niesler$^1$}
\address{
  $^1$Department of Electrical and Electronic Engineering, Stellenbosch University, South Africa\\
  $^2$CLS/CLST, Radboud University, Nijmegen, Netherlands\\
  $^3$Department of Electrical and Computer Engineering, National University of Singapore, Singapore\\
  $^4$UN Global Pulse, Kampala, Uganda \&
  $^5$University of Edinburgh, UK\vspace*{-1pt}}
\email{\footnotesize rmenon@sun.ac.za, kamperh@sun.ac.za, eleey@nus.edu.sg, john.quinn@unglobalpulse.org, trn@sun.ac.za\vspace*{-4pt}}

\usepackage[prependcaption,textsize=scriptsize]{todonotes}
\definecolor{mycolor}{HTML}{FF6600}

\begin{document}

\maketitle
\begin{abstract}
We consider multilingual bottleneck features (BNFs) for nearly zero-resource keyword spotting.  
This forms part of a United Nations effort using keyword spotting to support humanitarian relief programmes in parts of Africa where languages are severely under-resourced.
We use 1920 isolated keywords (40 types, 34 minutes) as exemplars for dynamic time warping (DTW) template matching, which is performed on a much larger body of untranscribed speech.
These DTW costs are used as targets for a convolutional neural network (CNN) keyword spotter, giving a much faster system than direct DTW.
Here we consider how available data from well-resourced languages can improve this CNN-DTW approach.
We show that multilingual BNFs trained on ten languages improve the area under the ROC curve of a CNN-DTW system by 10.9\% absolute relative to the MFCC baseline.
By combining low-resource DTW-based supervision with information from well-resourced languages, CNN-DTW is a competitive option for low-resource keyword spotting.

\end{abstract}

\section{Introduction}
Social media has become a popular medium for individuals to express opinions and concerns on issues impacting their
lives~\cite{Vosoughi_ICDMW15,Wegrzyn_CASoN11,Burnap15}. 
In countries without adequate internet infrastructure, like Uganda, communities often use phone-in talk shows on local radio stations for the same purpose.
In an ongoing project by the United Nations (UN), radio-browsing systems have been developed to monitor such radio shows~\cite{Menon2017,Saeb2017}.
These systems are actively and successfully supporting UN relief and developmental programmes.
The development of such systems, however, remains dependent on the availability of transcribed speech in the target languages.
This dependence has proved to be a key impediment to the rapid deployment 
of radio-browsing systems in new languages, since skilled annotators proficient in the target languages 
are hard to find, especially in crisis conditions.

In a conventional keyword spotting system, where the goal is to search through a speech collection for a specified set of keywords, automatic speech recognition (ASR) is typically used to generate lattices which are then searched to predict the presence or absence of keywords~\cite{Larson12, Mandal14}. 
State-of-the-art ASR, however, requires large amounts of transcribed speech audio~\cite{Sainath2015,Zhang2017}.
In this paper we consider the development of a keyword spotter without such substantial and carefully-prepared data. 
Instead, we rely only on a small number of isolated repetitions of keywords and a large body of untranscribed data from the target domain.
The motivation for this setting is that such isolated keywords should be easier to gather, even in a crisis scenario. 
 
Several studies have attempted ASR-free keyword spotting using a query-by-example (QbyE) retrieval procedure. 
In QbyE, the search query is provided as audio rather than text. 
Dynamic time warping (DTW) is typically used to search for instances of the query in a speech collection~\cite{Hazen2009, Zhang2009}.
As an alternative, several ways of obtaining fixed-dimensional representations of input speech have been considered~\cite{Levin2013}. 
Recurrent neural networks (RNNs)~\cite{Chen2015,settle+livescu_slt16}, autoencoding encoder-decoder RNNs~\cite{Chung2016,chung2018}, and Siamese convolutional neural networks (CNNs)~\cite{Kamper2016} have all been used to obtain such fixed-dimensional representations, which allow queries and search utterances to be directly compared without alignment.
For keyword spotting, a variant of this approach has been used where textual and acoustic inputs are mapped
into a shared 
space~\cite{Audhkhasi2017}.
Most of these neural approaches, however, relies on large amounts of training data.

In this paper, we extend the ASR-free keyword spotting approach first presented in~\cite{Menon2018}.
A small seed corpus of isolated spoken keywords is used to perform DTW template matching on a large corpus of untranscribed data from the target domain.
The resulting DTW scores are used as targets for 
training 
a CNN-based keyword spotter.
Hence we take advantage of DTW-based matching---which can be performed with limited data---and combine this with CNN-based searching---giving speed benefits since it does not require alignment.
In our previous work, we used speech data only from the target language. 
Here we consider whether data available for other (potentially well-resourced) languages can be used to improve performance.
Specifically, multilingual bottleneck features (BNFs) have been shown to provide improved performance by several authors~\cite{vesely2012language, Hermann2018, yuan2017pairwise, yuan2018}.
We investigate whether such multilingual bottleneck feature extractors (trained on completely different languages) can be used to extract features for our target data, thereby improving the overall performance of our CNN-DTW keyword spotting approach.

{
To perform a thorough analysis of our proposed approach (which requires transcriptions), we use a corpus of South African English.
We use BNFs trained on two languages and on ten languages as input features to the CNN-DTW system, and compare these to MFCCs.
We also consider features from unsupervised autoencoders, trained on unlabelled datasets from five languages.
We show that the 10-language BNFs work best overall, giving results that makes CNN-DTW a viable option for practical use.
}

\section{Radio browsing system}
\label{Sec_Radio}
\begin{figure}[t]
  \centering
  \captionsetup{justification=centering}
  \includegraphics[width=0.9\linewidth]{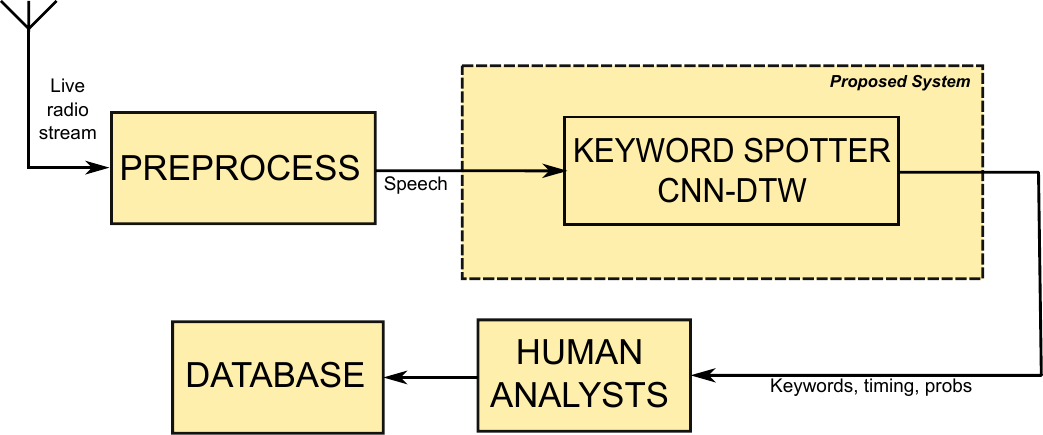}
  \vspace*{-5pt}
  \caption{Radio browsing using the CNN-DTW keyword spotter.}
  \label{fig:radio-browsing}
\end{figure}

The first radio browsing systems implemented as part of the UN's humanitarian monitoring programmes rely on ASR systems~\cite{Menon2017}. 
Human analysts filter speech segments identified by the system and add these to a searchable database to support decision making.\footnote{~\mbox{Examples at \url{http://radio.unglobalpulse.net}.}}
To develop the ASR system, at least a small amount of annotated speech in the target language is required~\cite{Saeb2017}.
However, the collection of even a small fully transcribed corpus has proven difficult or impossible in some settings.
In recent work, we have therefore proposed an ASR-free keyword spotting system based on CNNs~\cite{Menon2018}.
CNN classifiers typically require a large number of training examples, which are not available in our setting.
Instead, we therefore use a small set of recorded isolated keywords, which are then matched against a large collection of untranscribed speech drawn from the target domain using a DTW-based approach.
The resulting DTW scores are then used as targets for a CNN.
The key is that it is not necessary to know whether or not the keywords do in fact occur in this untranscribed corpus; the CNN is trained simply to emulate the behaviour of the DTW.
Since the CNN does not perform any alignment, it is computationally much more efficient than DTW.
The resulting CNN-DTW model can therefore be used to efficiently detect the presence of keywords in new input speech.
Figure~\ref{fig:radio-browsing} show the structure of this CNN-DTW radio browsing system.
 
\section{Data}
\label{Sec_Data}
We use the same datasets used in our previous work~\cite{Menon2018}.
As templates, we use a small corpus of isolated utterances of 40 keywords, each spoken twice by 24 South African speakers (12 male and 12 female).
This set of 1920 labelled isolated keywords constitutes the only transcribed target-domain data that we use to train our keyword spotter.
As untranscribed data, we use a corpus of South African Broadcast News (SABN). 
This 23-hour corpus consists of a mix of English newsreader speech, interviews, and crossings to reporters, broadcast between 1996 and 2006~\cite{Kamper2015}. 
The division of the corpus into different sets 
is shown in Table~\ref{SABC_data}.
The SABN training set is used as our untranscribed data to obtain targets for the CNN.
The models then perform keyword spotting on the SABN test set.
Since this set is fully transcribed, performance evaluation and analysis is possible.
The isolated keywords were recorded
under fairly quiet conditions and there was no speaker overlap with the SABN dataset.
Hence there is a definite mismatch between the datasets.
This is intentional as it reflects the intended operational setting of our system.

\begin{table}[!h]
\centering
\caption{The South African Broadcast News (SABN) dataset.}
\vspace*{-7.5pt}
\label{SABC_data}
{\eightpt
\renewcommand{\arraystretch}{1.2}
\begin{tabular}{|l|c|c|}
\hline
               & \textbf{Utterances} & \textbf{Speech (h)} \\ \hline\hline
{Train} & 5231                & 7.94                \\ 
{Dev}   & 2988                & 5.37                \\ 
{Test}  & 5226                & 10.33               \\ \hline
{Total} & 13445               & 23.64               \\ \hline
\end{tabular}}
\end{table}

\section{Keyword Spotting Approaches}
\label{sec:kyset}
Here we describe the combined CNN-DTW keyword spotting method. 
We also use direct DTW and a CNN classifier as baselines, and hence these are also briefly discussed.

\subsection{Dynamic time warping (DTW)}
\label{sec:dtw-ks}

In low-resource settings with scarce training data, DTW is an attractive approach, but it can be prohibitively slow since it requires repeated alignment. 
We make use of a simple DTW implementation in which isolated keywords slide over search audio, with a 3-frame-skip, and a frame-wise comparison is performed while warping the time axis. 
From this, a normalized per-frame cosine cost is obtained, resulting in a value $c \in [0, 2]$, with $0$ indicating a portion of speech that matches the keyword exactly. 
The presence or absence of the keyword is determined by applying an appropriate threshold to $c$.


\subsection{Convolutional neural network (CNN) classifier}
\label{sec:cnn}

As a baseline, we train a CNN classifier as an end-to-end keyword spotter. 
This would be typical in high resource settings~\cite{Kamper2016, palaz+etal_interspeech16, SainathPara2015}.  
We perform supervised training using the 1920 recorded isolated keywords with negative examples drawn randomly from utterances in the SABN training set.
For testing, a 60-frame window slides over the test utterances.
The presence or absence of keyword is again based on a threshold.\footnote{The CNN has 3 convolutional layers (filters with 64, 128, 256 units) with max pooling, followed by 3 dense layers (500, 100 and 300 neural units). We use a dropout of 0.5 for the first and last dense layer.}


\subsection{CNN-DTW keyword spotting}
\label{Sec:CNN-DTW}

\begin{figure}[t]
  \centering
	\captionsetup{justification=centering}
  \includegraphics[width=0.875\linewidth]{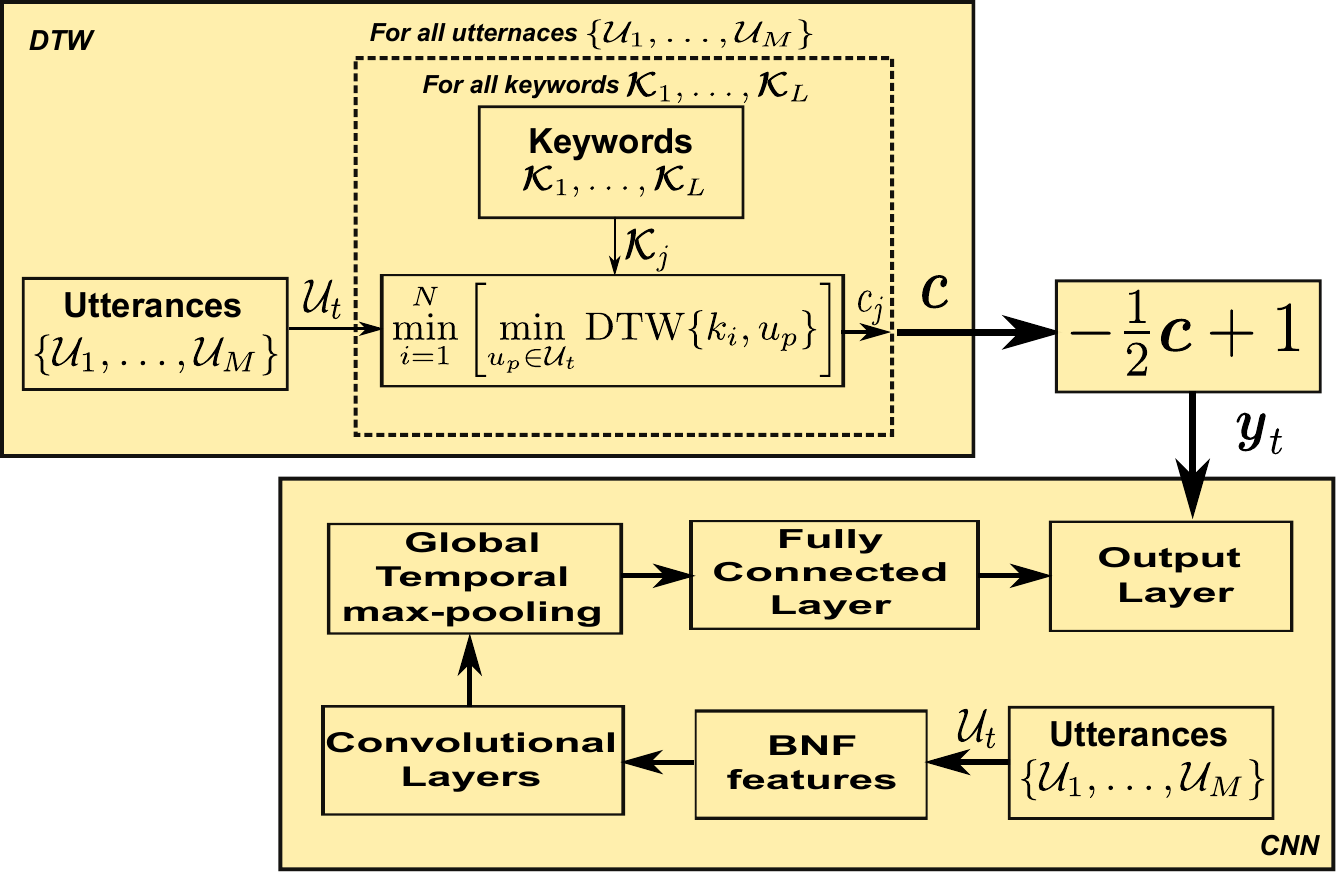}
  
  \vspace*{-5pt}
  \caption{The CNN-DTW keyword spotter using BNFs. The top half shows how the supervisory signal is obtained and the bottom half how this signal is used to train the CNN.}
   \vspace{-10pt}
  \label{fig:CNN_DTW}
\end{figure}

Rather than using labels (as in the CNN classifier above), the CNN-DTW keyword spotting approach uses DTW to generate sufficient training data as targets for a CNN.  
The CNN-DTW is subsequently employed as the keyword spotter; this is computationally much more efficient than direct DTW. 
DTW similarity scores are computed between our small set of isolated keywords and a much larger untranscribed dataset, and these scores are subsequently used as targets to train a CNN, as shown in Figure~\ref{fig:CNN_DTW}.
Our contribution here over our previous work~\cite{Menon2018} is to use multilingual BNFs instead of MFCCs, both for performing the DTW matching and as inputs to the CNN-DTW model. 
In Figure~\ref{fig:CNN_DTW}, the upper half shows how the supervisory signals are obtained using DTW, and the lower half shows how the CNN is trained.  
Equation~\eqref{eq:tar_vec} shows how keyword scores are computed, resulting in a vector $[ c_1, \ldots, c_j, \ldots c_L ]$ for each utterance $\mathcal{U}$.
\begin{equation}
    c = \min_{i \in 1 \ldots N} \left[ \min_{u_p \in \mathcal{U}} \textrm{DTW}\{k_{i},{u}_p\} \right]
    \label{eq:tar_vec}
  \end{equation}
Here, $k_{i}$ is the sequence of speech features for the $i^{th}$ exemplar of keyword $\boldsymbol{\mathcal{K}}$, ${u}_p$ is a successive segment of utterance $\mathcal{U}$, and $\textrm{DTW} \{ k_{i},{u}_p \}$ is the DTW alignment cost between the speech features of exemplar $k_i$ and the segment ${u}_p$. 
Each value $c_j$, which is between $[0, 2]$, is then mapped to $y_j \in [0, 1]$, with $1$ indicating a perfect match and $0$ indicating dissimilarity thus forming the target vector $\boldsymbol{y}$ for utterance $\mathcal{U}$. 
A CNN is then trained using a summed cross-entropy loss (which is why the scores are mapped to the interval $[0, 1]$) with utterance $\mathcal{U}$ as input and $\boldsymbol{y}$ as target. 
The CNN architecture is the same as used in~\cite{Menon2018}. 
Finally, the trained CNN is applied to unseen utterances.

\section{Bottleneck and Autoencoder Features}
\label{sec_BNF}
Our previous work focused purely on using data from the low-resource target language.
However, large annotated speech resources exist for several well-resourced languages.
We investigate whether such resources can be used to improve the CNN-DTW system in the unseen low-resource language.

\subsection{Bottleneck features}
One way to re-use information extracted from other multilingual corpora is to use multilingual bottleneck features (BNFs), which has shown to perform well in conventional ASR as well as intrinsic evaluations~\cite{vesely2012language,jcui2015,sthomas2012,Hermann2018,chen2017multilingual,yuan2017extracting}. 
These features are typically obtained by training a deep neural network jointly on several languages for which labelled data is available.
The bottom layers of the network are normally shared across all training languages.
The network then splits into separate parts for each of the languages, or has a single shared output. 
The final output layer has phone labels or HMM states as targets.
The final shared layer often has a lower dimensionality than the input layer, and is therefore referred to as a `bottleneck'.\footnote{Confusingly, the term `bottleneck' is now also sometimes used even when the layer does not have a smaller dimensionality than the input.}
The intuition is that this layer should capture aspects 
that are common across all the languages.
We use such features from a multilingual neural network in our CNN-DTW keyword spotting approach.
The BNFs are trained on a set of well-resourced languages different from the target language.




Different neural architectures can be used to obtain BNFs following the above methodology.
Here we use time-delay neural networks (TDNNs)~\cite{peddinti2015time}. 
We consider two models: a multilingual TDNN trained on only two languages, and a TDNN trained on ten diverse languages.
Our aim is to investigate whether it is necessary to have a large set of diverse languages, or whether it is sufficient to simply obtain features from a supervised model trained on a smaller set of languages.

\vspace{-7.5pt}
\subsubsection{2-language TDNN}
\vspace{-4pt}
\label{sec:2-lang}
An 11-layer 2-language TDNN was trained using 40-high resolution MFCC features as input on a combined set of Dutch and Frisian speech, as described in~\cite{yilmaz2018}.
Speaker adaptation is used with lattice-free maximum mutual information training, based on the Kaldi Switchboard recipe~\cite{povey2016}. 
Each layer uses ReLU activations with batch normalisation. 
By combining the the FAME~\cite{yilmaz2016} and CGN~\cite{cgn} corpora, the training set consists of a combined 
887 hours of data in the two languages.
40-dimensional BNFs are extracted from the resulting model.

\subsubsection{10-language TDNN}
\vspace{-4pt}
\label{sec:10-lang}
A 6-layer 10-language TDNN was trained on the GlobalPhone corpus, also using 40-high resolution MFCC features as input, as described in~\cite{Hermann2018}.
For speaker adaptation, a 100-dimensional i-vector was appended to the the MFCC input features. 
The TDNN was trained with a block-softmax, with the hidden layers shared across all languages and a separate output layer for each language. 
Each of the six hidden layers had 625 dimensions, and was followed by a 39-dimensional bottleneck layer with ReLU activations and batch normalisation. 
Training was accomplished using the Kaldi Babel receipe using 198 hours of data in 10 languages (Bulgarian, Czech, French, German, Korean, Polish, Portuguese, Russian, Thai, Vietnamese) from GlobalPhone.

\subsection{Autoencoder features}
\label{sec:ae}
BNFs are trained in a supervised fashion with acoustic feature presented at the input and phone targets at the outputs.
A more general scenario, however, is one in which training data is unlabelled, and these targets are therefore not known.
In this case, it may be possible to learn useful representations by using an unsupervised model. 
An autoencoder is a neural network trained to reconstruct its input.
By presenting the same data at the input and the output of the network while constraining intermediate connections, the network is trained to find an internal representation that is useful for reconstruction.
These internal representations can be useful as features~\cite{kamper2015unsupervised, badino2014auto,hinton2012deep,deng2010binary,sainath2012auto,gehring2013extracting}.
Like BNFs, autoencoders can be trained on languages different from the target language (often resulting in more data to train on).


Here we use a stacked denoising autoencoder~\cite{vincent2010stacked}. In this model, each layer is trained individually like an autoencoder with added noise to reconstruct the output of the previous layer.
Once a layer has been trained, its weights are fixed and its outputs  become the inputs to the next layer to be trained.
After all the layers are pre-trained in this fashion, the layers are stacked and fine-tuned. 
We use mean squared error loss and Adam optimisation~\cite{kingma2014adam} 
throughout.  We trained a 7-layer stacked denoising autoencoder on an untranscribed dataset consisting of 160 h of Acholi, 154 h of Luganda, 9.45 h of Lugbara, 7.82 h of Rutaroo and 18 h of Somali data. 
We used 39-dimensional MFCCs (13 cepstra with deltas and delta-deltas) as input and extracted features from the 39-dimensional fourth layer.

\section{Experimental setup}
\label{sec:Exp}

The experimental setup is similar to that of~\cite{Menon2018}.
We consider three baseline systems: two DTW systems and a conventional CNN classifier.
\begin{enumerate}
\setlength\itemsep{0em}
\item \textbf{DTW-QbyE}, where DTW is performed for each exemplar keyword on each utterance, and the resulting scores averaged (\S\ref{sec:dtw-ks}).
\item \textbf{DTW-KS}, where the minimum (best) score over all exemplars of a keyword type is used per utterance (\S\ref{sec:dtw-ks}). 
\item \textbf{CNN}, an end-to-end CNN classifier trained only on the isolated words (\S\ref{sec:cnn}).
\end{enumerate}
Our proposed approach, CNN-DTW, is supervised by the DTW-KS system.
Hyper-parameters for CNN-DTW were optimized using the target loss on the development set.\footnote{Final system: 10 convolutional layers (between 80 and 512 filters), two 3000-unit fully connected layers with a dropout of $0.5$, and a learning rate changing linearly from $10^{-4}$ to $10^{-5}$ used with Adam optimisation.}
Hence, the SABN transcriptions are not used for training or validation.
Performance is reported in terms of the area under the curve (AUC) of the receiver operating characteristic (ROC) and equal error rate (EER). 
The ROC is obtained by varying the detection threshold and plotting the false positive rate against the true positive rate. 
AUC, therefore, indicates the performance of the model independent of a threshold, with higher AUC indicating a better model. 
EER is the point at which the false positive rate equals the false negative rate and hence lower EER indicates a better model.

\section{Experimental results}

We consider four feature extractors in our experiments:
\begin{enumerate}
\setlength\itemsep{0em}
\item \textbf{SAE}, the stacked autoencoder (\S\ref{sec:ae}). 
\item \textbf{TDNN-BNF-2lang}, the 2-language TDNN without speaker normalisation (\S\ref{sec:2-lang}).
\item \textbf{TDNN-BNF-10lang}, the 10-language TDNN without speaker normalisation (\S\ref{sec:10-lang}).
\item \textbf{TDNN-BNF-10lang-SPN}, the 10-language TDNN with speaker normalisation (\S\ref{sec:10-lang}).
\end{enumerate}
In initial experiments, we first consider the performance of these features on development data.
Specifically, we use the features as representations in the DTW-based keyword spotter (DTW-KS).
Results are shown in Table~\ref{BNF_model_comp}.
BNFs trained on 10 languages outperform all other approaches, with speaker normalisation giving a further slight improvement.
Both the stacked autoencoder and the BNFs trained on two languages perform worse than the MFCC baseline.
This seems to indicate that a larger number of diverse languages is beneficial for training BNFs, and that supervised models are superior to unsupervised models when applied to an unseen target language.
However, further experiments are required to verify this definitively.
Based on these development experiments, we compare MFCCs and TDNN-BNF-10lang-SPN features when used for keyword spotting on evaluation data.


\begin{table}[t]
\centering
\captionsetup{justification=centering}
\caption{Performance of the different features on the development set when used in a DTW-based keyword spotter.}
\label{BNF_model_comp}
\vspace{-7.5pt}
{\eightpt
\renewcommand{\arraystretch}{1.2}
\begin{tabular}{|l|l|l|}
\hline
\multicolumn{1}{|c|}{\multirow{2}{*}{\textbf{Model}}} & \multicolumn{2}{c|}{\textit{\textit{\textbf{dev}}}}                            \\ \cline{2-3} 
\multicolumn{1}{|c|}{}                                & \multicolumn{1}{c|}{\textbf{AUC}} & \multicolumn{1}{c|}{\textbf{EER}} \\ \hline \hline
{MFCC}                                & 0.7556                            & 0.3092                            \\ 
{SAE}                                 & 0.5247                            & 0.4844                            \\ 
{TDNN-BNF-2lang}                    & 0.7273                            & 0.3356                            \\ 
{TDNN-BNF-10lang}                   & 0.7725                            & 0.2884                            \\ 
{TDNN-BNF-10lang-SPN}              & \textbf{0.7781}                            & \textbf{0.2872}                            \\ \hline
\end{tabular}}
\vspace{-10pt}
\end{table}

Table~\ref{table_Performance} shows the performance of the three baseline systems and CNN-DTW when using MFCCs and BNFs. 
In all cases except the CNN classifier, BNFs lead to improvements over MFCCs. 
Furthermore, we see that, when using BNFs, the CNN-DTW system performs almost as well as its DTW-KS counterpart.
The DTW-KS system provided the targets with which the CNN-DTW system was trained, and hence represents an upper bound on the performance we can expect from the CNN-DTW wordspotter.
When using BNFs, we see that the difference between the DTW-KS and CNN-DTW approaches becomes smaller compared to the difference for MFCCs.
This results in the CNN-DTW system using BNFs almost achieving the performance of the DTW-KS system; the former, however, is computationally much more efficient since alignment is not required.
On a conventional desktop PC with a single NVIDIA GeForce GTX 1080 GPU, CNN-DTW takes approximately 5 minutes compared to DTW-KS which takes 900 minutes on a 20-core CPU machine.
Table~\ref{table_Performance} shows that, in contrast to when MFCCs are used, a Gaussian noise layer (CNN-DTW-GNL) does not give further performance benefits for the BNF systems.

\begin{figure}[!t]
\centering
\captionsetup{justification=centering}
\subfloat[Part 1][\vspace{-4pt}Keyword: Government]{\includegraphics[width=1.5in]{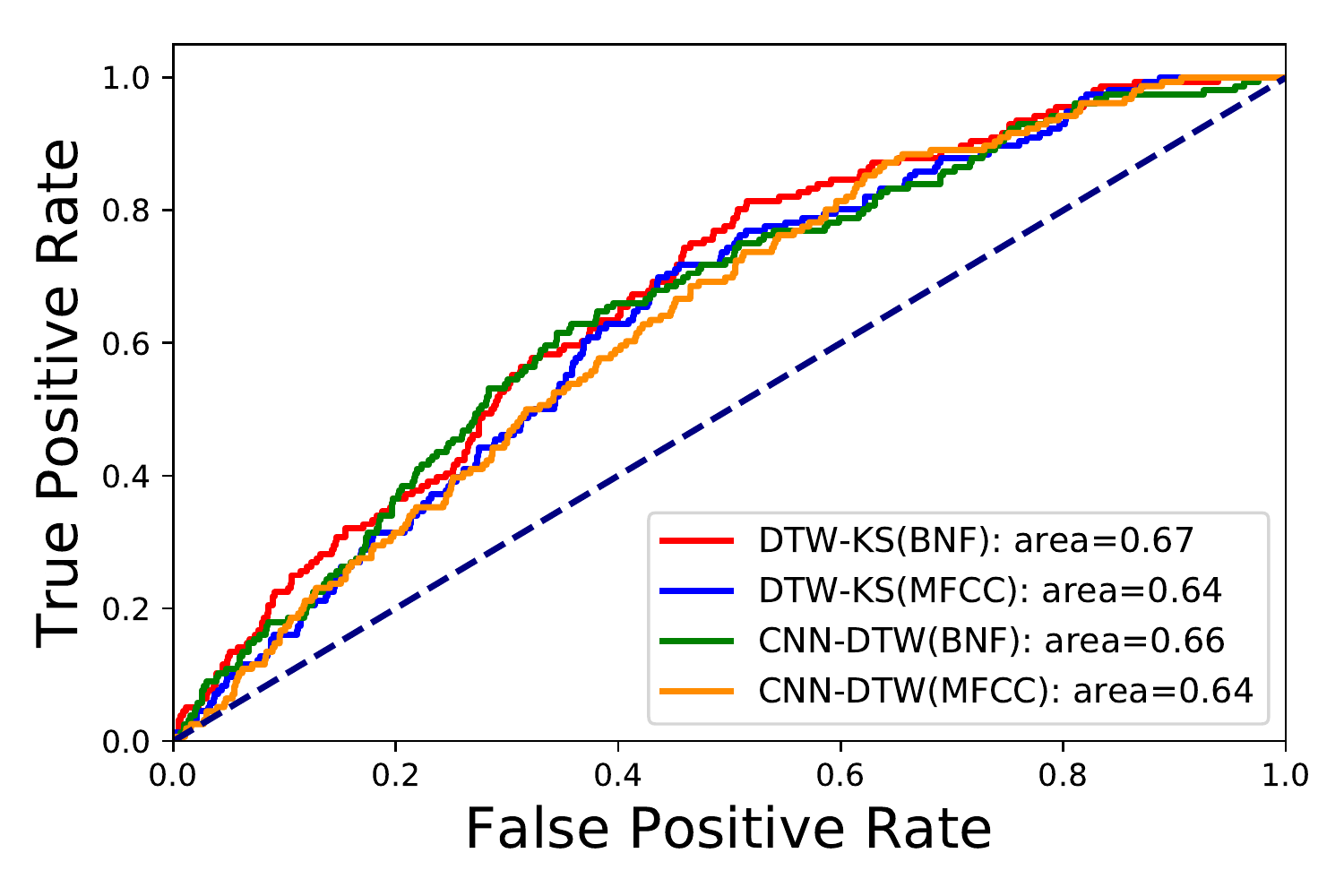} \label{fig:roc_a}}
\subfloat[Part 2][\vspace{-4pt}Keyword: Attack]{\includegraphics[width=1.5in]{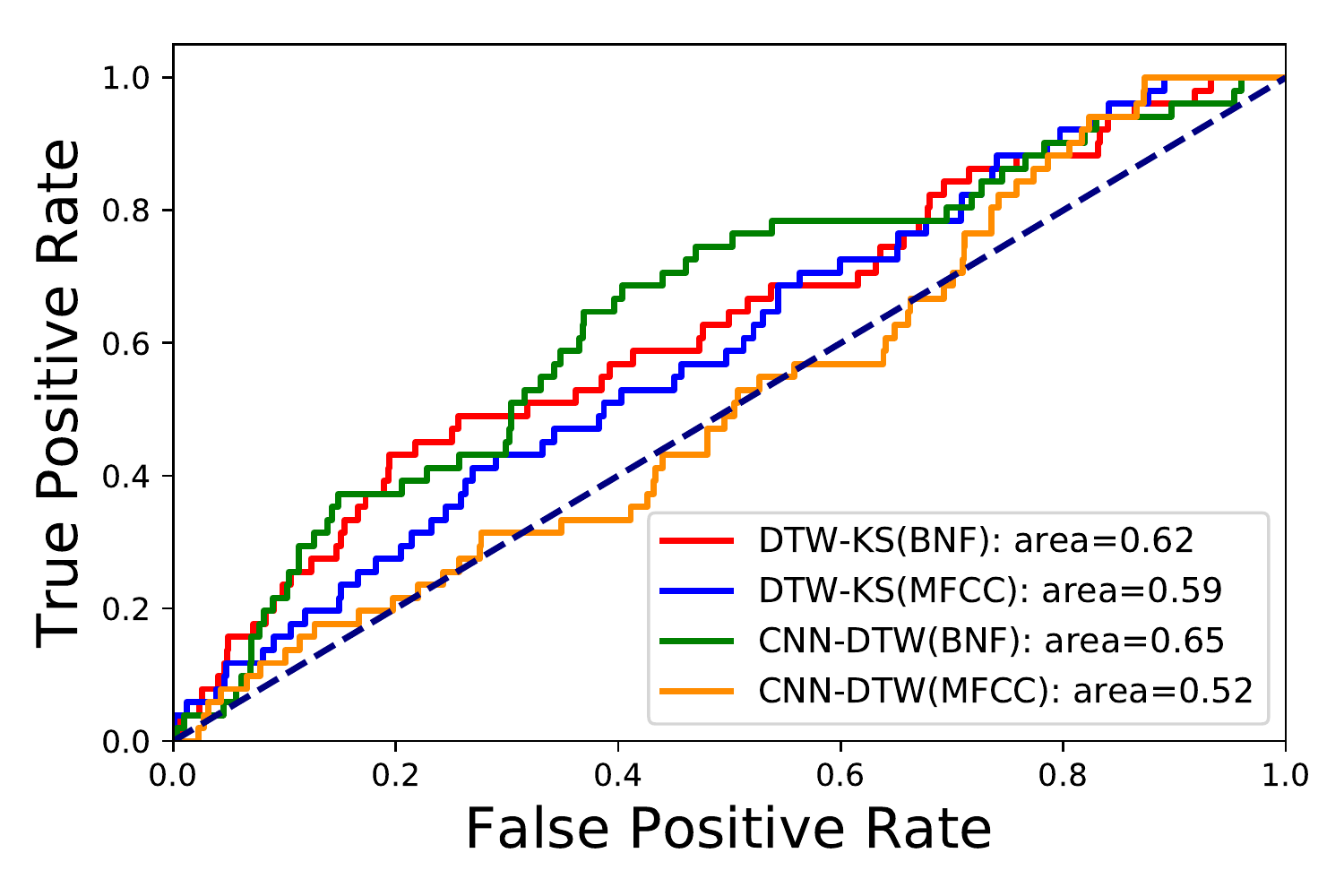} \label{fig:roc_b}}\\[-6pt]
\subfloat[Part 3][\vspace{-4pt}Keyword: HIV]{\includegraphics[width=1.5in]{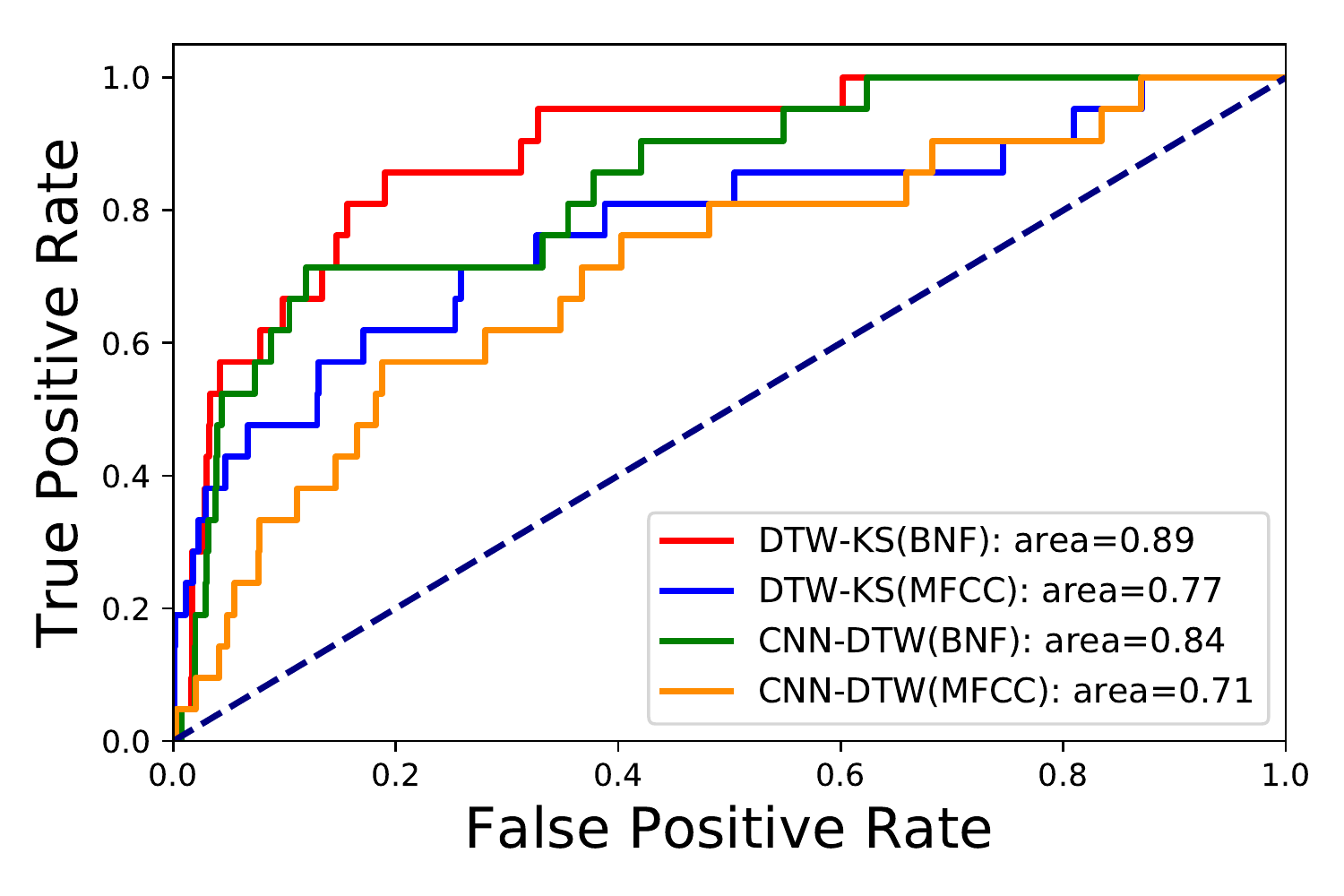} \label{fig:roc_c}}
\subfloat[Part 4][\vspace{-4pt}Keyword: Health]{\includegraphics[width=1.5in]{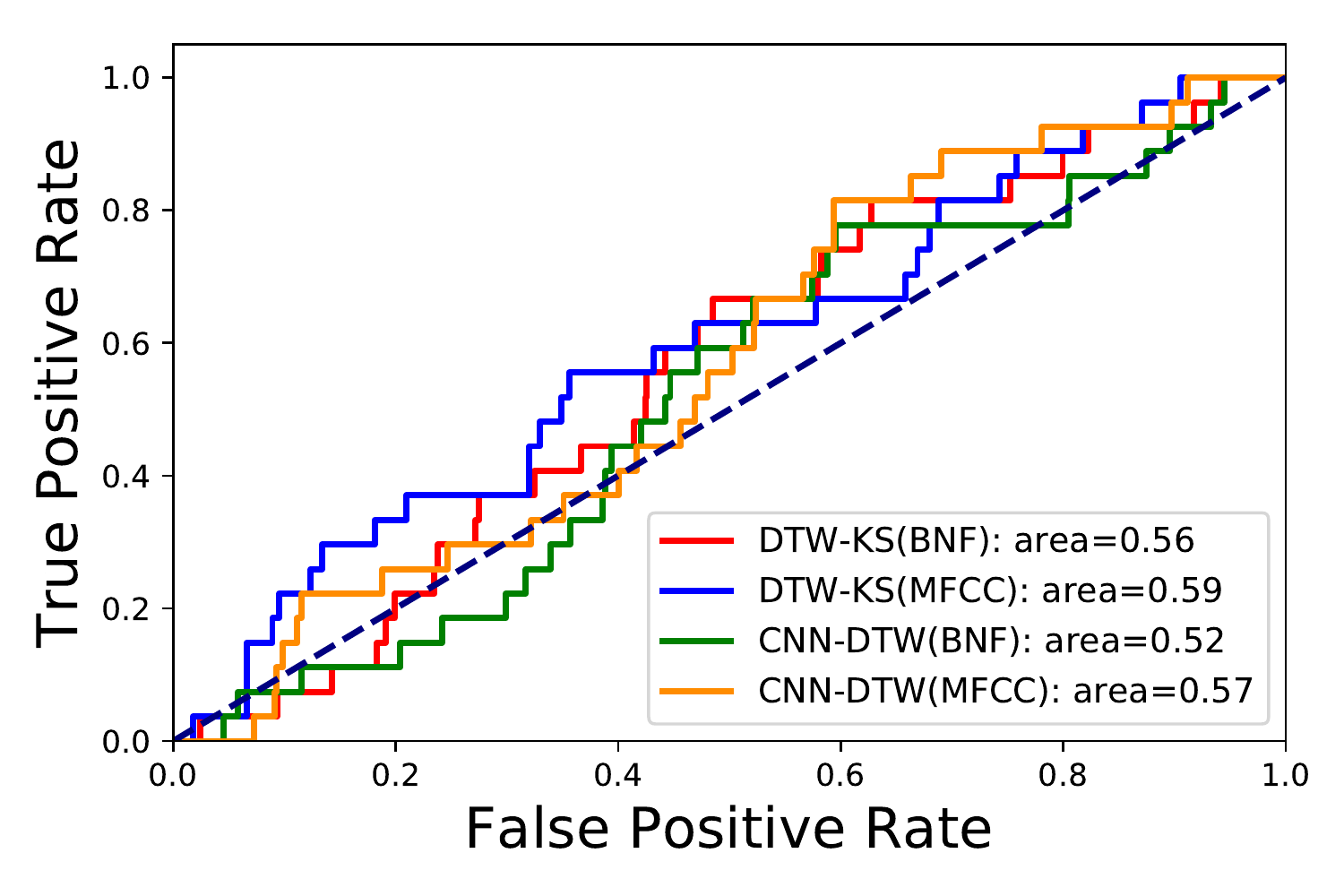} \label{fig:roc_d}}\\[-6pt]
\subfloat[Part 5][\vspace{-4pt}Keyword: War]{\includegraphics[width=1.5in]{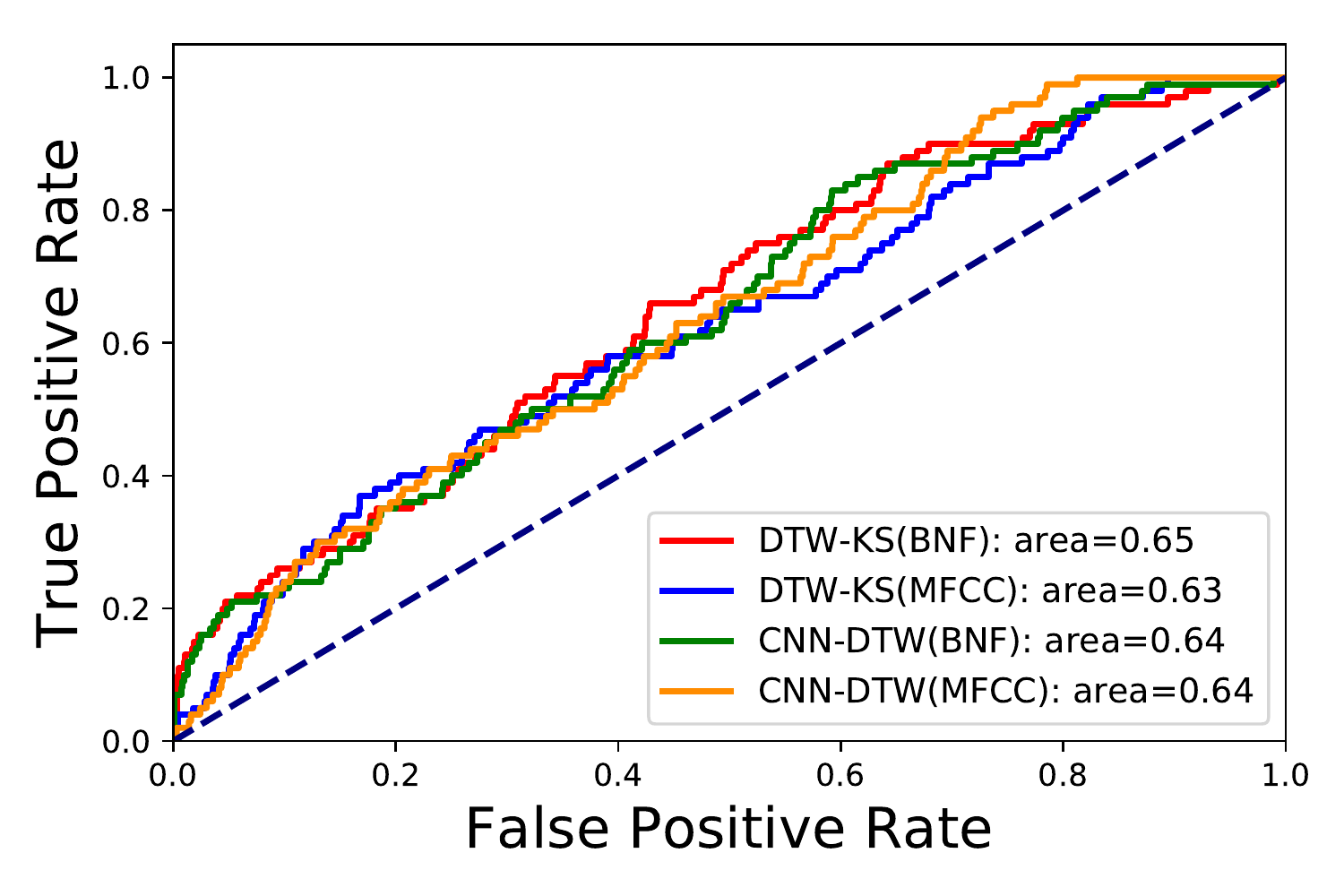} \label{fig:roc_e}}
\subfloat[Part 6][\vspace{-4pt}Keyword: Wounded]{\includegraphics[width=1.5in]{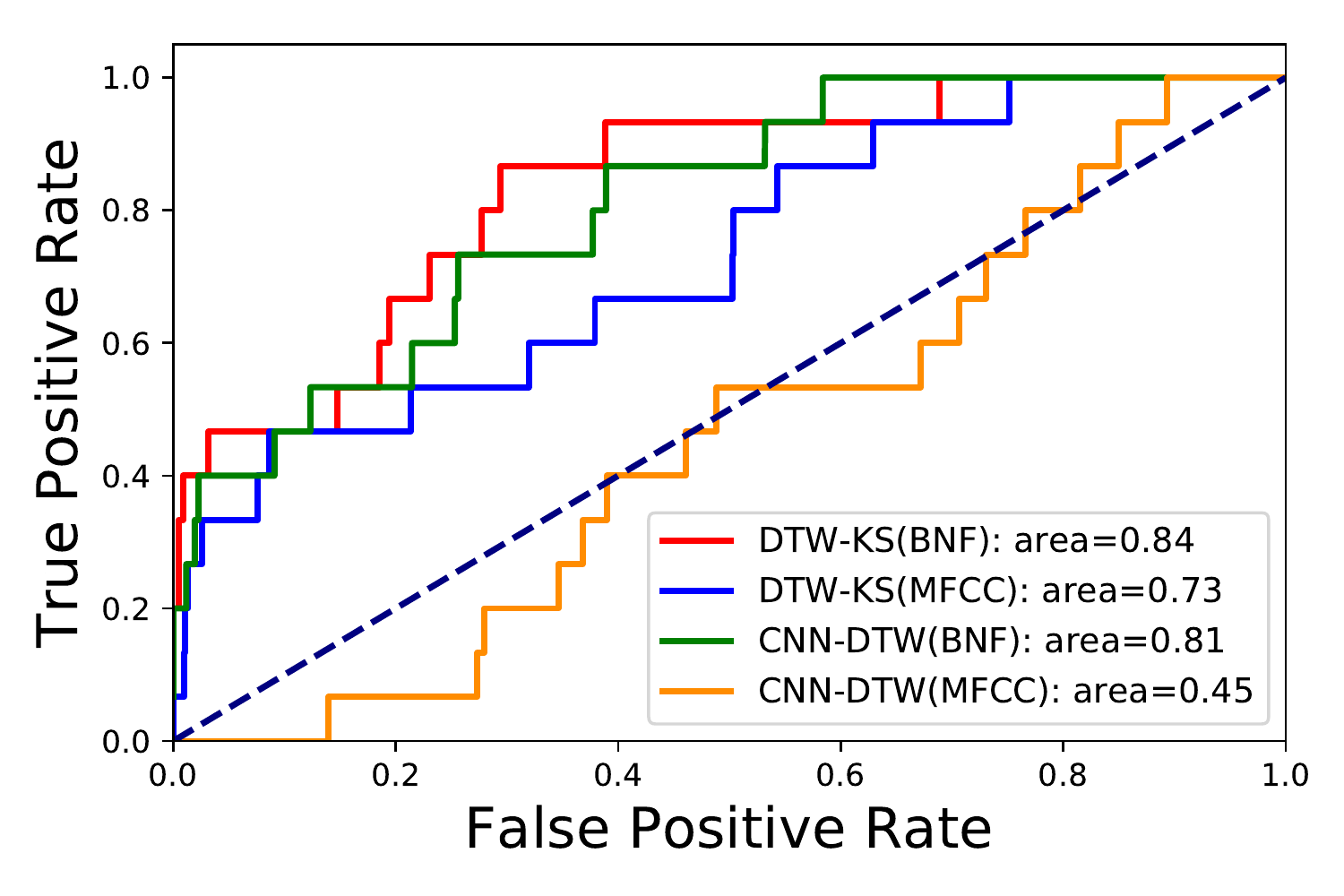} \label{fig:roc_f}}
\vspace{-1pt}
\caption{Receiver operating characteristic curves for selected keywords for the DTW keyword spotter and the CNN-DTW system when using MFCCs and BNFs.}
\label{fig:ROC}
\vspace*{-10pt}
\end{figure}

Figures~\ref{fig:ROC}(a-f) show ROC plots for a selection of keywords 
which are representative of cases with both good and bad performance. 
AUC improves in all cases when switching from MFCCs to BNFs, except for \textit{health}, where the difference is relative small (all scores are close to chance on this keyword).
In some cases, e.g.\ for \textit{wounded}, the benefits of switching to BNFs in CNN-DTW is substantial.
Interestingly, for keywords such as \textit{attack}, the CNN-DTW system using BNFs actually marginally outperforms the DTW-KS system which is used to supervise it.


\begin{table}[t]
\centering
\captionsetup{justification=centering}
\caption{Performance of different keyword spotting systems using MFCCs and BNFs (TDNN-BNF-10lang-SPN).}
\label{table_Performance}
\vspace{-7.5pt}
{\eightpt
\renewcommand{\arraystretch}{1.3}
\resizebox{\columnwidth}{!}{%
\begin{tabular}{|l|l|l|l|l|l|l|l|l|}
\hline
\multicolumn{1}{|c|}{\multirow{3}{*}{\textbf{Model}}} & \multicolumn{4}{c|}{\textbf{AUC}} & \multicolumn{4}{c|}{\textbf{EER}} \\ \cline{2-9} 
\multicolumn{1}{|c|}{} & \multicolumn{2}{c|}{\textit{\textbf{dev}}} & \multicolumn{2}{c|}{\textit{\textbf{test}}} & \multicolumn{2}{c|}{\textit{\textbf{dev}}} & \multicolumn{2}{c|}{\textit{\textbf{test}}} \\ \cline{2-9} 
 & \multicolumn{1}{c|}{\textbf{MFCC}} & \multicolumn{1}{c|}{\textbf{BNF}} & \multicolumn{1}{c|}{\textbf{MFCC}} & \multicolumn{1}{c|}{\textbf{BNF}} & \multicolumn{1}{c|}{\textbf{MFCC}} & \multicolumn{1}{c|}{\textbf{BNF}} & \multicolumn{1}{c|}{\textbf{MFCC}} & \multicolumn{1}{c|}{\textbf{BNF}} \\ \hline \hline
{{CNN}}                                                        & 0.5698 & 0.5298 & 0.5448 & 0.5364 & 0.4435 & 0.4813 & 0.4771 & 0.4725 \\ 
{{DTW-QbyE}}                                                   & 0.6639 & 0.6899 & 0.6612 & 0.6873 & 0.3864 & 0.3556 & 0.3885 & 0.3661 \\ 
{{DTW-KS}}                                                     & 0.7556 & 0.7781 & 0.7515 & 0.7699 & 0.3092 & 0.2872 & 0.3162 & 0.3012 \\ 
{{CNN-DTW}}                                                    & 0.6360 & 0.7537 & 0.6285 & 0.7422 & 0.4073 & 0.3058 & 0.4161 & 0.3214 \\ 
{{CNN-DTW-GNL}}                                                & 0.6443 & 0.7535 & 0.6357 & 0.7518 & 0.4036 & 0.3091 & 0.4092 & 0.3153 \\ \hline
\end{tabular}%
}}
\vspace{-10pt}
\end{table}

\section{Conclusion}
\label{sec:Con}
We investigated the use of multilingual bottleneck (BNFs) and autoencoder features in a CNN-DTW keyword spotter.
While autoencoder features and BNFs trained on two languages did not improve performance over MFCCs, BNFs trained on a corpus of 10 languages lead to substantial improvements.
We conclude that our overall CNN-DTW based approach, which combines the low-resource advantages of DTW with the speed advantages of CNNs, further benefits by incorporating labelled data from well-resourced languages through the use of BNFs when these are obtained from several diverse language.

\vspace*{\itemsep}\noindent
\textbf{Acknowledgements:} We thank the NVIDIA corporation for the donation of GPU equipment used for this research. We also gratefully acknowledge the support of Telkom South Africa.

\bibliographystyle{IEEEtran}

\bibliography{mybib}


\end{document}